\documentclass[conference]{IEEEtran}
\IEEEoverridecommandlockouts
\usepackage{cite}
\usepackage{amsmath,amssymb,amsfonts}
\usepackage{graphicx}
\usepackage{textcomp}
\usepackage{xcolor}
\def\BibTeX{{\rm B\kern-.05em{\sc i\kern-.025em b}\kern-.08em
    T\kern-.1667em\lower.7ex\hbox{E}\kern-.125emX}}

\usepackage{pdflscape}
\usepackage{verbatim}
\usepackage{eurosym}

\usepackage{graphicx}
\usepackage{multirow}
\usepackage{pifont}

\usepackage{hyperref}

\usepackage{listings}
\usepackage{framed}
\usepackage[listings,skins]{tcolorbox}
\usepackage[skipbelow=\topskip,skipabove=\topskip]{mdframed}
\mdfsetup{roundcorner=0}
\usepackage{etoolbox}

\usepackage{algorithmicx}
\usepackage{algorithm}
\usepackage{algpseudocode}
\algnewcommand\algorithmicinput{\textbf{Input:}}
\algnewcommand\INPUT{\item[\algorithmicinput]}

\algnewcommand\algorithmicoutput{\textbf{Output:}}
\algnewcommand\OUTPUT{\item[\algorithmicoutput]}

\AtBeginDocument{}

\usepackage{url}

\begin{document}

\title{Exploring Opportunistic Meta-knowledge to Reduce Search Spaces for Automated Machine Learning}

\author{\IEEEauthorblockN{Tien-Dung Nguyen, David Jacob Kedziora, Katarzyna Musial, Bogdan Gabrys}
\IEEEauthorblockA{Complex Adaptive Systems Lab, School of Computer Science \\
University of Technology Sydney, Australia \\
TienDung.Nguyen-2@student.uts.edu.au, \\ \{David.Kedziora, Katarzyna.Musial-Gabrys, Bogdan.Gabrys\}@uts.edu.au}
}



\maketitle

\begin{abstract}

Machine learning (ML) pipeline composition and optimisation have been studied to seek multi-stage ML models, i.e. preprocessor-inclusive, that are both valid and well-performing.
These processes typically require the design and traversal of complex configuration spaces consisting of not just individual ML components and their hyperparameters, but also higher-level pipeline structures that link these components together. 
Optimisation efficiency and resulting ML-model accuracy both suffer if this pipeline search space is unwieldy and excessively large; it becomes an appealing notion to avoid costly evaluations of poorly performing ML components ahead of time.
Accordingly, this paper investigates whether, based on previous experience, a pool of available classifiers/regressors can be preemptively culled ahead of initiating a pipeline composition/optimisation process for a new ML problem, i.e. dataset.
The previous experience comes in the form of classifier/regressor accuracy rankings derived, with loose assumptions, from a substantial but non-exhaustive number of pipeline evaluations; this meta-knowledge is considered `opportunistic'.
Numerous experiments with the AutoWeka4MCPS package, including ones leveraging similarities between datasets via the relative landmarking method, show that, despite its seeming unreliability, opportunistic meta-knowledge can improve ML outcomes.
However, results also indicate that the culling of classifiers/regressors should not be too severe either. In effect, it is better to search through a `top tier' of recommended predictors than to pin hopes onto one previously supreme performer.

\end{abstract}

\section{Introduction}
\label{sec:intro}

Various ML pipeline composition and optimisation methods have been proposed to construct valid and well-performing multi-stage ML models, given both a problem (i.e. a dataset) and a set of ML components with tunable hyperparameters \cite{sabu18,zohu19,ngma20,kemu20}. Typically, this pool of ML components contains classification/regression predictors and other preprocessing operators, e.g. for imputation or feature generation/selection.
Among ML pipeline composition and optimisation methods, one of the most successful is based on the Sequential Model-based Algorithm Configuration (SMAC) approach \cite{thhu13}. Like most optimisers, this method seeks a balance between the exploration and exploitation of configuration spaces. When exploiting, the procedure investigates ML pipelines that are similar to the current best performers in terms of pipeline structure and hyperparameter values. When exploring, the procedure selects random candidates within configuration space instead, seeking to avoid entrapment in any local optima.

There are several automated machine learning (AutoML) tools that implement SMAC, starting with AutoWeka version 0.5 \cite{thhu13}. Most of these seek one-component pipelines and thus search through configuration spaces that only involve predictors and their hyperparameters. AutoWeka4MCPS is a rare exception, both extending configuration spaces for data preprocessing components and upgrading the implementation of SMAC, thus enabling the construction and optimisation of multiple-component pipelines \cite{sabu18}.
While this extension of configuration space allows a wider range of diverse and possibly better ML solutions to be explored, it does come with a number of challenges.
Key among them is that a large configuration space is more difficult to efficiently traverse for any optimiser. Given that every candidate ML pipeline must also be trained/queried on a dataset to evaluate its accuracy, and that training can be computationally expensive, this can be a substantial obstacle for using AutoML on a novel ML problem.
There is therefore both a need for and a great interest in approaches that offer the intelligent reduction of configuration spaces by preemptively excluding unpromising ML pipelines or, more severely, ML components. 

Several previous attempts have been made to deal with the problem of configuration space reduction, although few have considered the additional intricacies involving pipeline structure. These approaches typically lean one of two ways when culling the search space: hard restrictions defined by expert knowledge \cite{sabu18,fekl15,depi17,tsga12} and dynamic constraints based on meta-learning \cite{olmo16,wemo18,giya18,dbbr18,va19,lebu15,albu15,rihu18,prbo19,wemu20}.
The latter notion is of particular appeal to AutoML as it is effectively hands-off; the solution to a new ML problem is aided by the automatic extraction of `meta-knowledge' from previous experience.
Problematically, though, advocacy of meta-learning often hinges on the curation of an ideal meta-knowledge base, which, in the AutoML context, would need to involve intensive evaluations of many ML components, each one thoroughly sampled across a frequently multi-dimensional range of hyperparameter values. To make matters more complicated, the performance of ML components can vary substantially between two intrinsically dissimilar datasets, and it is not even clear what kind of dataset characteristics should be a metric for that dissimilarity \cite{lega10,lega10a,albu15}.
In practice, data scientists do not have access to a tailor-made meta-knowledge base. On the other hand, in the natural course of executing AutoML pipeline composition/optimisation processes on a dataset, data scientists do implicitly acquire accuracy evaluations for numerous ML-pipeline candidates. So, we ask the question: are these evaluations opportunistically useful? Can they recommend
how many and which ML components we should select when designing an ML-pipeline search space? 

To explore these research questions, we run a series of experiments with the AutoWeka4MCPS package \cite{sabu18}, which is accelerated by the ML-pipeline validity checker, AVATAR \cite{ngga20}, wherever specified. All experiments revolve around a meta-knowledge base that is built by using loose assumptions to convert limited SMAC-based AutoML runs across 20 datasets into mean-error statistics and associated performance rankings for 30 Weka predictors, both overall and per dataset. The meta-knowledge base is considered to be neither rigorous nor exhaustive. Despite this, the experiments seek to address whether rankings from the compiled statistics are still reliable enough to guide an improved search for an ML solution to a new problem.
Some experiments additionally explore whether this meta-knowledge can be improved by considering dataset similarity; in these cases, we employ the relative landmarking method \cite{va19} to quantify that similarity.
Because SMAC has not completed the ML pipeline composition and optimisation tasks for 9 datasets for at least one configuration space, we only present experiments on 11 datasets.

Ultimately, the main contributions of this study are: 

\begin{itemize}
\item An investigation of how the performance of an AutoML composition/optimisation process is affected by varying levels of recommended pipeline search space reduction, i.e. removing all but the `best' $k$ of 30 predictors from an ML-component pool for variable $k$.
\item An exploration of how those results vary under different modes of recommendation, e.g. the best predictors over all datasets versus the best predictors for the most similar dataset, all derived from opportunistic and somewhat unreliable meta-knowledge.
\end{itemize}

    


Accordingly, this paper is divided into five sections. After the Introduction,
Section \ref{sec:related_work} reviews previous attempts to reduce configuration spaces in the context of AutoML.
Section \ref{sec:methodology} details the methodology used in this study, e.g. dynamic configuration spaces, meta-knowledge generation, and relative landmarking.
Section \ref{sec:experiment} presents and analyses experiments assessing whether meta-learned recommendations for culling configuration space are beneficial to the performance of pipeline composition/optimisation. Finally, Section \ref{sec:conclusion} concludes this study.

\section{Related Work}
\label{sec:related_work}

The growing number of available ML methods with their often complex hyperparameters leads to a very rapid expansion, if not combinatorial explosion, of ML-pipeline configurations and associated search spaces. Intelligent reduction of these configuration spaces enables ML pipeline composition and optimisation methods to find valid and well-performing ML pipelines faster within the typical constraints of execution environments and time budgets.
We review two main approaches to reduce configuration spaces in the context of ML pipeline composition and optimisation. 

\textit{Predefined ML pipeline structures and component hyperparameters:} This approach can be implemented as fixed pipeline templates \cite{sabu18,fekl15,depi17,tsga12} or ad-hoc specifications \cite{olmo16,wemo18,giya18,dbbr18,va19,lebu15,albu15,rihu18,prbo19,wemu20} such as context-free grammars. Moreover, specific ranges of hyperparameter values, which highly contribute to well-performing pipelines, are also predefined in these specifications.
The advantage of this approach is its simple nature, leveraging expert knowledge to reduce configuration spaces by directly restricting the length of ML pipelines, the pool of ML components, and their permissible orderings/arrangements. However, the disadvantage of this approach is that expert bias might obscure strongly performing ML pipelines outside of the predefined templates.

\textit{Meta-learning:} This approach aims to reduce configuration spaces by using prior knowledge to avoid wasting time with unpromising ML-solution candidates. Frequently, this involves assessing similarity between past and present ML problems/datasets, so as to hone in on the most relevant meta-knowledge available \cite{lega10,albu15,lebu15,dbbr18,va19}. 
To quantify this similarity, characteristics are typically established for datasets, which can then be used in correlations. A characteristic can be directly derived from the dataset as a meta-feature, e.g. the number of raw features or data instances. Alternatively, relevant to this study, two datasets can be compared by the relative performance of landmarkers. These landmarkers are ideally simple one-component pipelines, i.e. predictors, that are of varying types; they estimate the suitability of varying modelling approaches for a dataset. For instance, the performance of a linear regressor theoretically quantifies whether an ML problem is linear. An ML problem that is estimated to be nonlinear will likely not benefit from methods serving a linear dataset. In any case, the meta-learning approach can be used to reduce configuration spaces by selecting a number of well-performing ML components  \cite{albu15,dbbr18,va19,lebu15} or important hyperparameters for tuning \cite{albu15,rihu18,prbo19,wemu20}. For instance, both \textit{average ranking} and \textit{active testing} have previously been used to recommend ML solutions for new datasets \cite{dbbr18}. However, these approaches have not been applied to AutoML yet. Moreover, these studies limit their scopes by optimising predictors, not multi-component pipelines, and the optimisation method they use is grid search, proven not to be as effective as SMAC \cite{thhu13}. 
Other studies have investigated estimating the importance of hyperparameters \cite{rihu18,prbo19,wemu20} from prior evaluations.
Specifically, some hyperparameters are more sensitive to perturbation than others; tuning them can contribute to a proportionally higher variability in ML-algorithm performance, i.e. error rate. As an example, gamma and complexity variable C are the most important hyperparameters for a support vector machine (SVM) with RBF kernel \cite{rihu18}.
Consequently, the results of these studies can be used to reduce configuration spaces by constraining less-important hyperparameters, either by making their search ranges less granular or outright fixing them as default values.  
This frees up more time to seek the best values for important hyperparameters that have the highest impact on finding well-performing pipelines.
However, a disadvantage of these studies is that the importance of ML-component hyperparameters has only been studied on small sets of up to six algorithms. This reflects how time-consuming it is to properly sample hyperparameter space across all available algorithms. 

In this study, our approach aligns with meta-learning principles. However, it differs from previous research by refusing to carefully curate a tailor-made meta-knowledge base. Instead, accepting a degree of unreliability, we opportunistically derive assumptive statistics from numerous pipeline evaluations; these evaluations are non-exhaustive and simulate the remnants of AutoML optimisation processes intended to solve seemingly unrelated problems. 
Accepting this context, we identify previously well-performing ML components, sometimes weighted by the past-and-present dataset similarity derived via the relative landmarking method, and we constrain configuration subspaces for ML pipeline composition and optimisation around these top performers. We also investigate how varying degrees of this recommendation-based search-space culling affects the performance of ML pipeline composition/optimisation methods.


\section{Meta-learning Methodology for Configuration Space Reduction}
\label{sec:methodology}

Here, we present the methods used in the three major facets of our meta-learning study. Section \ref{sec:problem_formulation} describes how pipeline configuration space is designed and augmented for dynamic re-sizing. Section \ref{sec:meta_construction} details how we construct a meta-knowledge base, acknowledging its intentional limitations. Section \ref{sec:landmarkers} covers the specifics of relative landmarking and its use in identifying similar datasets.

\subsection{Dynamic Configuration Spaces}
\label{sec:problem_formulation}
Broadly stated, an ML model is a mathematical object that attempts to approximate a desired function. It is typically paired with an ML algorithm that, via the process of training, feeds on encountered data to adjust certain variables, i.e. model parameters, so as to improve the accuracy of the approximation. This pairing of ML model and algorithm, an ML component, contains other variables, i.e. hyperparameters, that are fixed throughout the training process. Hyperparameter optimisation is thus the process of finding values for these training constants that optimise the performance of the trained model, usually via some iterative approach. Even at this level, the task is not trivial; hyperparameter space can involve many dimensions that are continuous or discrete, with varying ranges.

When hyperparameter optimisation extends to variable ML components, a core facet of AutoML, configuration space becomes even more complex, involving so-called `conditional' hyperparameters. For instance, the polynomial degree of an SVM kernel is only non-null if the type of SVM kernel is set to polynomial. Consequently, the search space for a single-component model is better represented by a tree structure, which SMAC is suited to handle \cite{thhu13}.

The incorporation of pipeline structure in AutoML search space complicates matters. It has been done before \cite{sabu18}, but our study necessitated an auxiliary representation of pipelines, specified as paths through a tree-structured space of ML components. This allows nodes to be marked active/inactive at any time so that an augmented SMAC can include/avoid ML pipelines containing those components while searching through that space. In effect, ML components can be pruned from configuration space to leave a substantially smaller `active' subspace.

For the sake of brevity, we defer presenting any mathematical formalism until a later work. However, by taking the above approach, we reframe the equations for pipeline search \cite{sabu18} as a constrained optimisation problem. Moreover, while our investigation only ever culls the search space once per dataset, per experiment, our methodology is suited to more dynamic modulations of search space; we leave this to future research.

\subsection{The Meta-knowledge Base}
\label{sec:meta_construction}

Prior to any meta-learning experiments and analysis, a meta-knowledge base must first be constructed. However, to simulate the desired `coincidental' nature of the metadata and its availability, we limit the collection of previous experience to SMAC-based AutoML applied across 20 datasets, and only a singular two-hour run per dataset at that. This is enough to generate numerous pipeline evaluations, extracted from iterations of the optimising algorithm SMAC, but it still falls far short of the exhaustive exploration that a meta-knowledge base ideally requires. This is especially true, as a single evaluation does not just fix pipeline structure, it also fixes values for a set of hyperparameters. In fact, to be technical, a single SMAC iteration is one-tenth of a 10-fold cross-validatory ML-pipeline evaluation; given enough time, up to ten SMAC iterations can be dedicated to the same pipeline/hyperparameter configuration. Moreover, per dataset, a single optimisation path is a very poor sampling of an entire configuration space. Some ML components may feature negligibly in the evaluated pipelines, if at all. Time budgets also complicate matters; some ML solutions may be more computationally expensive to train than others, and some SMAC runs may, via exploration/exploitation, end up in these regions of configuration space, leading to an unbalanced distribution of evaluations across datasets. In essence, the quality of accumulated experience is expected to be highly variable.

Another issue is that, in raw form, pipeline evaluations are relatively useless; any one instantiation, hyperparameter sampling included, is unlikely to be visited again by SMAC in the future. Generalisations must thus be made if configuration space is to be effectively reduced. To that end, we make a loose assumption that, in the absence of further information, the error of a pipeline represents a sampled error of its constituent predictor. From this, mean-error statistics and associated performance rankings can be compiled for 30 Weka predictors, both overall and per dataset. These are much more practical, as a subspace forged around $k$ out of 30 predictors is a much more substantial reduction than excluding individual pipelines. Of course, the assumption behind the generalisation is very contentious, as the selection of preprocessors in a pipeline will obviously affect the accuracy of its predictor.

So, intentionally working with limited meta-knowledge and presumptuous generalisations, the question is: are the compiled statistics still useful for narrowing in on promising subspaces?




\subsection{Landmarkers}
\label{sec:landmarkers}

Typical reasoning in the field of meta-learning is that previous experience is most relevant to a problem at hand if past and present contexts are similar. Accordingly, it is routine to approach this by defining and compiling a set of so-called meta-features to describe a dataset, which are then subsequently compared between datasets. Naturally, identifying the most appropriate metrics to denote this similarity is a topic of active research, but landmarking has proved to be a popular option \cite{va19}; we employ this procedure in relevant experiments.

A set of landmarkers, $\Theta=\{\theta_i\}$, is generally a collection of ML predictors that are simple and efficient to execute. Ideally, they represent a diversity of problem types. The theory is that, if a landmarker is well-suited for problem type $A$, and it produces an ML model with strong performance, e.g. good classification accuracy, on dataset $B$, then dataset $B$ belongs to the class of problems designated by $A$. Any ML pipeline that works well for one dataset in class $A$ is then presumed to work well for any other of that same problem type.
However, in practice, it is challenging to pick a perfect set of landmarkers, especially as the choice of meta-features to describe complex problems has an impact on the effectiveness of similarity-based meta-learning \cite{va19,albu15}. Given that we include the evaluation of landmarkers as part of the overall AutoML optimisation time within relevant experiments, we have made a deliberate decision in this study to prioritise fast execution time. Therefore, sourced from the average evaluation time of all predictor-containing pipelines in our meta-knowledge base, we select the following five fastest predictors for our set of landmarkers: RandomTree, ZeroR, IBk, NaiveBayes, and OneR. We acknowledge that this choice is relatively crude, but it adheres to the opportunistic principles behind this study; are rough metrics for dataset similarity still useful in providing additional intelligence when reducing the input search space for AutoML pipeline selection?


\begin{algorithm}[!htbp]
\small

\caption{\small Designing Configuration Space with Relative Landmarking}
\begin{algorithmic}[1]
\Require
      \Statex $\Theta$: The set of landmarkers
      \Statex $t_{new}$: The new dataset
      \Statex \{$t_{prior_j}$\}: The set of prior datasets
\Statex      
\For{$\theta_i$ \textbf{in}  $\Theta$}
        \State $E_{new\_i}$ = \textit{evaluate}($\theta_i$, $t_{new}$)
\EndFor
\For{\textbf{each}  $t_{prior\_j}$}
        \State $c_j$ = 
        \textit{calculateCorrelation}($E_{new}$, $E_{prior\_j}$)
\EndFor
\State $t^{*}$ = \textit{getMostSimilarTask}($c$) 
\State $\mathcal{T}_{new}$ = \textit{selectKBestMLComponents}($t^{*}$, $k$)
\State \textbf{return} $\mathcal{T}_{new}$
\end{algorithmic}
\label{algorithm:construct_configuration_space}
\end{algorithm}

Algorithm \ref{algorithm:construct_configuration_space} formalises how configuration space is constrained via the relative landmarking method, to then be used as input for AutoML pipeline composition and optimisation methods. Firstly, the algorithm evaluates the new dataset $t_{new}$ with each landmarker $\theta_i$, resulting in a 10-fold cross-validation error rate, $E_{new\_i}$, per landmarker (lines 1-3). Secondly, the algorithm calculates a Pearson correlation coefficient between the full performance vector of the new dataset, $E_{new}$, and a similarly landmarked vector of mean error rates, $E_{prior\_j}$, for each prior dataset $t_{prior\_j}$ (lines 4-6). Thirdly, the algorithm ranks the correlation coefficients and selects the dataset, $t^{*}$, that has the highest correlation coefficient (line 7). Finally, the resulting configuration space to explore is constructed from all preprocessing components and the \textit{k} best performing predictors (line 8) for the most similar dataset, $t^{*}$. We emphasise that, for landmarker-based experiments on a newly encountered dataset, the net evaluation time of landmarkers is deducted from the total time budget assigned to ML pipeline composition/optimisation processes.

\section{Experiments}
\label{sec:experiment}

To explore the opportunistic utility of a limited meta-knowledge base, constructed according to Section \ref{sec:meta_construction}, we run a series of experiments with different settings, all described in Section \ref{sec:settings}. The results are described in Section \ref{sec:results}.

\subsection{Experimental settings}
\label{sec:settings}

All experiments take the form of running AutoWeka4MCPS\footnote{https://github.com/UTS-AAi/autoweka} across 20 datasets listed in Table \ref{tab:datasets}. The AutoML package uses SMAC \cite{thhu13} for ML pipeline composition/optimisation and is applied for two hours of runtime and 1 GB of memory per dataset, although this two-hour process is itself done five times over for statistical purposes. 
Logged results typically note the 10-fold cross-validated error rate of the best ML pipeline across the five repeated runs. 
Essentially, the experiments only vary in how the searchable configuration space has been recommended to the AutoML package, i.e. which $k$ out of 30 predictors to utilise.

First, we define the `control' contexts, in which meta-learning does not feature:
\begin{itemize}
    \item \textbf{baseline}: The full configuration space constructed from all preprocessing and predictor components available to AutoWeka4MCPS \cite{sabu18}.
    
    \item \textbf{avatar}: The full configuration space as per \textbf{baseline}, but where the AutoML solution search process is significantly boosted by AVATAR \cite{ngma20}, identifying and ignoring invalidly composed ML pipelines before they can waste evaluation time.
    
    \item \textbf{r30}: An extreme case, where the pipeline structure of an ML solution, per dataset, is fixed to the best that was found after a previous 30 hours optimisation of AutoML as reported in \cite{sabu18}. The two hours in this `continuation' experiment are solely dedicated to optimising hyperparameters that have been re-initialised to their default values.
    

\end{itemize}
    
Next, we define the contexts that pull information from the meta-knowledge base, noting that the AVATAR speed-up is employed for all:
\begin{itemize}
    \item \textbf{global leaderboard (\textit{M-k1}, \textit{M-k4}, \textit{M-k8}, \textit{M-k10}, and \textit{M-k19})}: Untargeted meta-knowledge. For each dataset, AutoML explores pipelines containing $k$ predictors, for $k$ in $\{1,4,8,10,19\}$, that performed the best across all datasets. This global leaderboard is constructed by averaging the rank numbers of a predictor from each individual dataset, then sorting these averages for all predictors.

    \item \textbf{landmarked (\textit{L-k1}, \textit{L-k4}, \textit{L-k8}, \textit{L-k10}, and \textit{L-k19})}: Targeted meta-knowledge. For each dataset, AutoML explores pipelines containing $k$ predictors, for $k$ in $\{1,4,8,10,19\}$, that performed the best on the most similar dataset, where similarity is defined by landmarkers; see Section \ref{sec:landmarkers}.

    \item \textbf{oracle (\textit{O-k1}, \textit{O-k4}, \textit{O-k8}, \textit{O-k10}, and \textit{O-k19})}: A representation of direct memory. For each dataset, AutoML explores pipelines containing $k$ predictors, for $k$ in $\{1,4,8,10,19\}$, that performed the best on the same dataset, according to previous runs in the meta-knowledge base.

    
\end{itemize}


Notably, we restrict the set of $k$ values for the meta-learning experiments due to the computational expense in running them. The particular spread of $\{1,4,8,10,19\}$ is the consequence of an initial exploratory experiment, 
which we exclude detailing here for the sake of brevity. As a result, these numbers may seem unusual, but they have no grander significance beyond being 
one possible way to sample the broad spectrum of predictor-culling scenarios.

\begin{table}[htb!]
\centering
\caption {Dataset characteristics: the number of numeric attributes, nominal attributes, the number of distinct classes, instances in training and testing sets.}
\label{tab:datasets}
\fontsize{7.2}{8.5}\selectfont
\begin{tabular}{|l|l|l|l|l|l|}
\hline
\textbf{Dataset}  & \textbf{Numeric} & \textbf{Nominal} & \textbf{Distinct classes} & \textbf{Train} & \textbf{Test} \\ \hline \hline
abalone           & 7                & 1                & 28                              & 2,924          & 1,253         \\ \hline
car               & 0                & 6                & 4                               & 1,210          & 518           \\ \hline
dexter            & 20,000           & 0                & 2                               & 420            & 180           \\ \hline
gcredit           & 7                & 13               & 2                               & 700            & 300           \\ \hline
krvskp            & 0                & 36               & 2                               & 2,238          & 958           \\ \hline
madelon           & 500              & 0                & 2                               & 1,820          & 780           \\ \hline
semeion           & 256              & 0                & 10                              & 1,116          & 477           \\ \hline
shuttle           & 9                & 0                & 7                               & 43,500         & 14,500        \\ \hline
waveform          & 40               & 0                & 3                               & 3,500          & 1,500         \\ \hline
winequality       & 11               & 0                & 11                              & 3,429          & 1,469         \\ \hline
yeast             & 8                & 0                & 10                              & 1,039          & 445           \\ \hline \hline
adult             & 6                & 8                & 2                               & 32,561         & 16,281        \\ \hline
amazon            & 10,000           & 0                & 50                              & 1,050          & 450           \\ \hline
cifar10small      & 3,072            & 0                & 10                              & 10,000         & 10,000        \\ \hline
convex            & 784              & 0                & 2                               & 8,000          & 50,000        \\ \hline
dorothea          & 100,000          & 0                & 2                               & 805            & 345           \\ \hline
gisette           & 5,000            & 0                & 2                               & 4,900          & 2,100         \\ \hline
kddcup & 192              & 38               & 2                               & 35,000         & 15,000        \\ \hline
mnist             & 784              & 0                & 10                              & 12,000         & 50,000        \\ \hline
secom             & 590              & 0                & 2                               & 1,097          & 470           \\ \hline

\end{tabular}

\end{table}


In any case, conventional wisdom and the no-free-lunch theorem suggest that, controlling for variability, \textit{baseline} should result in the worst model performance, with \textit{avatar} being an improvement due to an effective decrease in runtime wastage. Scenario \textit{r30} is difficult to predict \textit{a priori}, but, with a 30-hour head-start on optimising pipeline structure, it should outperform \textit{baseline} as well. As for the meta-learning experiments, they are ordered in increasing relevance of meta-knowledge to the dataset/problem on hand; the solutions found by AutoML should improve along this ordering. In effect, for $k=n$ and with $E(x)$ representing the cross-validated error of the optimal ML pipeline, we would naively expect
\begin{eqnarray}
\label{eq:hierarchy}
    \nonumber && E(\mathrm{baseline})>E(\mathrm{r30}),\\
    \nonumber && E(\mathrm{baseline})>E(\mathrm{avatar}),\\
    && E(\mathrm{avatar})>E(\operatorname{M-kn})>E(\operatorname{L-kn})>E(\operatorname{O-kn}).
\end{eqnarray}
Accordingly, the results, and any deviation from expectation, are analysed with Eq. \ref{eq:hierarchy} in mind.

\subsection{Experiment Results}
\label{sec:results}

In the course of running the 18 experimental settings detailed in Section~\ref{sec:settings}, SMAC encountered 9 datasets for which its optimisation process was, for at least one configuration space, unable to evaluate any ML pipeline the 10 times required by cross-validation. While these `incomplete' runs are still worthy of discussion in future work, all results presented in this section relate to the remaining datasets, i.e. the first 11 in Table~\ref{tab:datasets}.


\begin{table*}[htb!]

\vspace{-0.35cm}

\caption {Mean error rate (\%) of the pipelines found by SMAC using different configuration spaces.   }
\label{tab:error_rate_all_methods}

\centering
\setlength\tabcolsep{3.0pt}
\begin{tabular}{|l|l|l|l|l|l|l|l|l|l|l|l|l|l|l|l|l|l|l|}
\hline
\textbf{Dataset} & \textbf{baseline} & \textbf{r30}  & \textbf{avatar} & \textbf{M-k1}  & \textbf{M-k4}  & \textbf{M-k8} & \textbf{M-k10} & \textbf{M-k19} & \textbf{L-k1}  & \textbf{L-k4} & \textbf{L-k8} & \textbf{L-k10} & \textbf{L-k19} & \textbf{O-k1}  & \textbf{O-k4}  & \textbf{O-k8} & \textbf{O-k10} & \textbf{O-k19} \\ \hline 
abalone          & 73.46             & 73.81         & 73.85           & \textbf{72.27} & 72.41          & 72.89         & 73.26          & 73.18          & 72.74          & 72.95         & 73.40         & 72.98          & 73.27          & 73.77          & 73.42          & 73.31         & 73.28          & 72.93          \\ \hline
car              & 2.73              & 0.35          & 0.58            & 5.92           & 2.98           & 2.71          & 2.93           & 0.38           & \textbf{0.33}  & 0.35          & \textbf{0.33} & \textbf{0.33}  & 0.38           & 2.60           & 0.35           & 0.36          & \textbf{0.33}  & 0.55           \\ \hline
dexter           & 8.81              & 8.21          & 8.10            & 8.45           & 9.95           & 10.71         & 9.57           & 8.76           & \textbf{5.33}  & 7.19          & 8.05          & 7.90           & 8.95           & 6.29           & 5.67           & 7.80          & 8.39           & 8.93           \\ \hline
gcredit          & 22.83             & 23.26         & 22.26           & 21.74          & 22.14          & 22.57         & 22.63          & 22.23          & 23.49          & 23.00         & 23.00         & 22.57          & 22.14          & 23.00          & \textbf{21.23} & 21.49         & 21.91          & 21.80          \\ \hline
krvskp           & 0.67              & \textbf{0.38} & 0.44            & 0.44           & 0.40           & 0.45          & 0.40           & 0.48           & 0.78           & 0.74          & 0.40          & 0.40           & 0.43           & 0.55           & 0.41           & 0.42          & 0.44           & 0.46           \\ \hline
madelon          & 26.04             & 39.41         & 22.86           & 26.63          & \textbf{22.64} & 22.89         & 23.02          & 23.78          & 33.84          & 22.66         & 22.68         & 22.68          & 22.88          & 22.66          & 22.72          & 23.04         & 22.91          & 23.41          \\ \hline
semeion          & 8.37              & 5.47          & 4.95            & 5.95           & 6.00           & 7.97          & 7.44           & 6.01           & 4.46           & \textbf{4.45} & 4.53          & 4.59           & 5.36           & 8.65           & 4.77           & 5.41          & 5.52           & 6.93           \\ \hline
shuttle          & 0.031             & 0.022         & 0.087           & 0.017          & \textbf{0.017} & 0.022         & 0.020          & 0.020          & 0.253          & 0.043         & 0.028         & 0.025          & 0.037          & \textbf{0.017} & 0.019          & 0.018         & 0.019          & 0.026          \\ \hline
waveform         & 12.71             & 12.90         & 12.53           & 13.83          & 13.95          & 12.54         & 12.55          & 12.57          & \textbf{12.44} & 12.55         & 12.46         & 12.47          & 12.58          & 12.57          & 12.50          & 12.57         & 12.54          & 12.49          \\ \hline
winequality      & 37.84             & 34.57         & 33.52           & \textbf{32.72} & 32.81          & 33.84         & 33.50          & 33.39          & 39.01          & 33.21         & 33.56         & 33.45          & 33.92          & \textbf{32.72} & 32.91          & 32.89         & 33.29          & 35.19          \\ \hline
yeast            & 39.15             & 39.94         & 38.02           & 36.75          & 36.94          & 37.71         & 37.65          & 38.00          & 41.20          & 38.52         & 37.80         & 37.94          & 37.80          & \textbf{36.65} & 36.90          & 37.63         & 37.42          & 37.55          \\ \hline
\end{tabular}

\end{table*}

\begin{figure*}[htbp]
\centering
\includegraphics[width=0.95\linewidth]{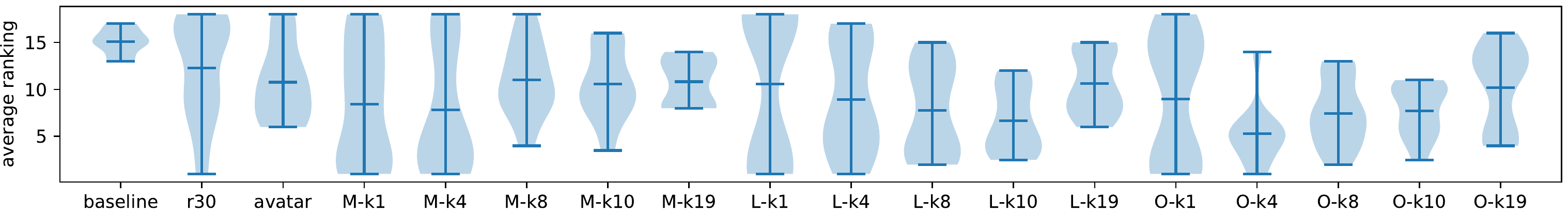}
\caption{The violin plots capturing the average rankings and ranking distribution for the configuration spaces on 11 datasets.}
\label{fig:violin_chart}
\end{figure*}

\begin{table}[htb!]
\centering
\vspace{-0.55cm}
\caption {The best ML pipelines found using different configuration spaces for the dataset \textit{abalone}.}
\label{tab:pipelines}
\fontsize{7.2}{8.5}\selectfont
\setlength\tabcolsep{3.2pt}
\begin{tabular}{|l|l|}
\hline
\textbf{Configuration space} & \textbf{Pipeline structure}                                                                                                                                             \\ \hline 
baseline                     & SimpleLogistic                                                                                                                                                          \\ \hline
r30                          & \begin{tabular}[c]{@{}l@{}}CustomReplaceMissingValues\\      $\rightarrow$ RandomSubset $\rightarrow$ Resample\\      $\rightarrow$ Logistic $\rightarrow$ Bagging\end{tabular}                     \\ \hline
avatar                       & SMO                                                                                                                                                                     \\ \hline
L-k1                         & \begin{tabular}[c]{@{}l@{}}CustomReplaceMissingValues\\      $\rightarrow$ Normalize $\rightarrow$ RandomSubset $\rightarrow$ SimpleLogistic\end{tabular}                                    \\ \hline
L-k4                         & \begin{tabular}[c]{@{}l@{}}ClassBalancer $\rightarrow$ RemoveOutliers $\rightarrow$ InterquartileRange\\      $\rightarrow$ Normalize $\rightarrow$ RandomSubset $\rightarrow$ SimpleLogistic\end{tabular} \\ \hline
L-k8                         & REPTree                                                                                                                                                                 \\ \hline
L-k10                        & RandomForest                                                                                                                                                            \\ \hline
L-k19                        & PART                                                                                                                                                                    \\ \hline
O-k1                         & Logistic                                                                                                                                                                \\ \hline
O-k4                         & DecisionTable                                                                                                                                                           \\ \hline
O-k8                         & SimpleLogistic                                                                                                                                                          \\ \hline
O-k10                        & PART                                                                                                                                                                    \\ \hline
O-k19                        & MultilayerPerceptron                                                                                                                                                    \\ \hline
M-k1                         & RandomForest                                                                                                                                                            \\ \hline
M-k4                         & RandomForest                                                                                                                                                            \\ \hline
M-k8                         & PART                                                                                                                                                                    \\ \hline
M-k10                        & REPTree                                                                                                                                                                 \\ \hline
M-k19                        & MultilayerPerceptron                                                                                                                                                    \\ \hline
\end{tabular}

\end{table}



\begin{figure}[h] 
    \centering
    \includegraphics[width=0.90\linewidth]{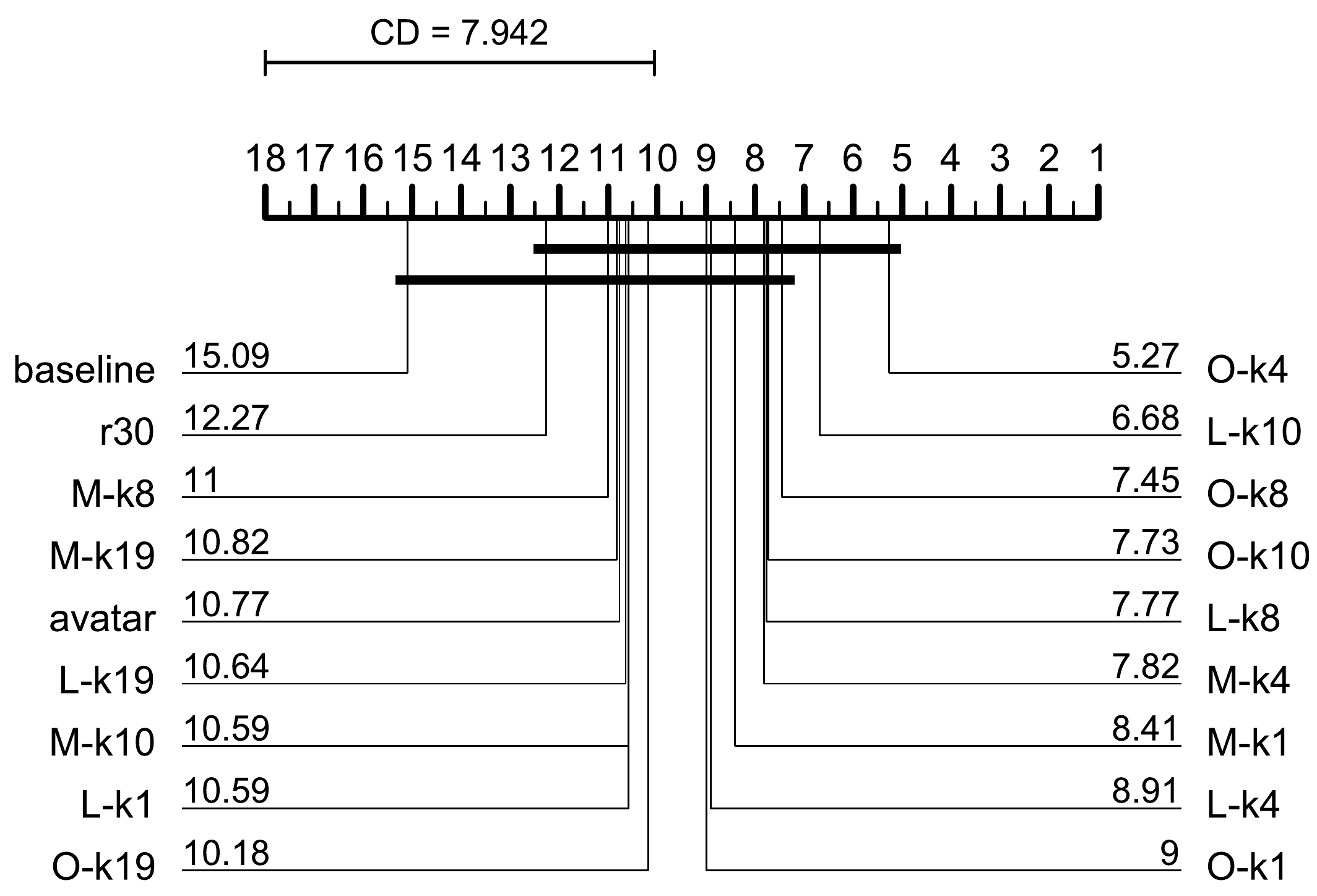}
    \vspace*{-1mm}
    \caption{The critical difference diagram of the average ranking of the performance of SMAC with different configuration spaces.}
    \label{fig:cd_diagram}
\end{figure}

With this acknowledged, Table \ref{tab:error_rate_all_methods} presents the mean error rate (\%) of the best pipelines found by SMAC for each different method of designing a searchable configuration space; the lowest value per dataset is shown in bold. Table \ref{tab:pipelines}\footnote{The details of the best ML pipelines found by SMAC with different configuration spaces for all data sets can be found at \url{https://github.com/UTS-AAi/autoweka/blob/master/autoweka4mcps/doc/landmarking_supplementary.pdf} } demonstrates what these optimal ML pipelines look like for the specific dataset \textit{abalone}. As for the performance measures, they allow experimental settings to be ranked in utility from 1 to 18 for each dataset. Equal rankings are averaged out, e.g. the top four settings for dataset \textit{car} are all ranked $(1+2+3+4)/4=2.5$.
Immediately, given the rankings for each configuration scenario, the distribution across all 11 datasets can be displayed in Fig. \ref{fig:violin_chart}. A comparison of their averages is likewise depicted in
Fig. \ref{fig:cd_diagram} via a critical difference (CD) diagram.

\begin{figure}[htbp]
\includegraphics[width=0.90\linewidth]{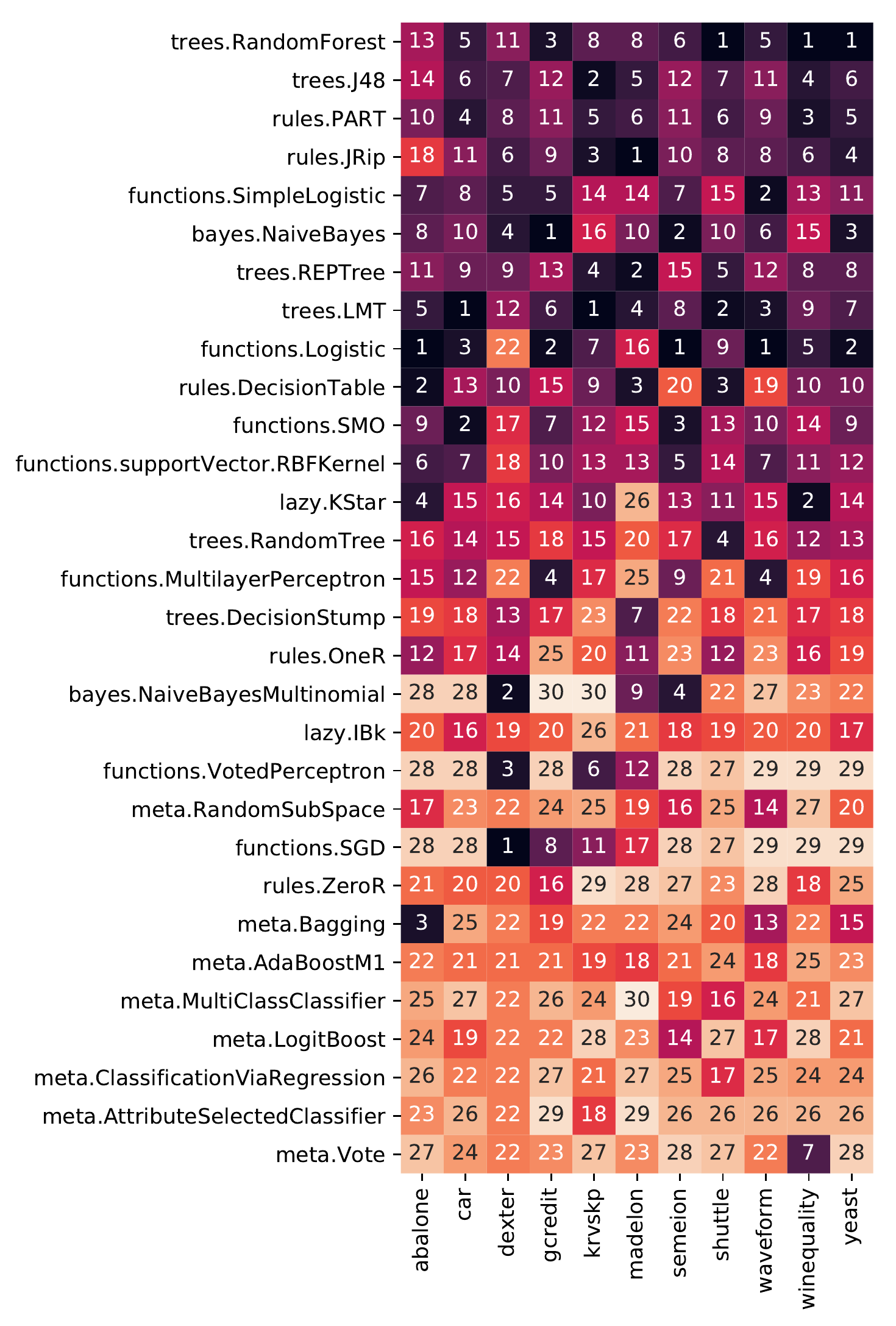}
\vspace*{-3mm}
\caption{The ranking of ML predictor components based on mean error rate of their pipelines from prior evaluations for selected datasets.}
\label{fig:ranking_predictors}
\end{figure}

Reassuringly, the expectation in Eq. (\ref{eq:hierarchy}) is upheld, if loosely, in both the CD diagram and the ranking distributions. Crudely averaging the averages already presented in Fig.~\ref{fig:cd_diagram} provides the following mean rankings: 9.728 for \textit{M-kn}, 8.918 for \textit{L-kn}, and 7.926 for \textit{O-kn}. When assessed against values of 15.09 for \textit{baseline}, improved to 12.27 for \textit{r30} and 10.77 for \textit{avatar}, the benefits of meta-learning seem both evident and additive with respect to refined targeting.
Indeed, using \textit{global leaderboard} configuration spaces is a decent strategy, likely to yield `good enough' performance for the majority of datasets, with \textit{M-k4} and \textit{M-k1} proving quite a bit better than the \textit{baseline}, \textit{r30} and \textit{avatar} scenarios.
The \textit{landmarked} configuration spaces are an upgrade beyond the \textit{global leaderboard}, using dataset similarity to select the best ML components that are, hopefully, relevant to a problem on hand; sure enough, \textit{L-k10} and \textit{L-k8} appear better than \textit{M-k4} and \textit{M-k1}.
Finally, \textit{oracle} configuration spaces ideally trump all, leveraging the best predictors found previously for each dataset on the same dataset. It is no surprise then that \textit{O-k4}, \textit{O-k8} and \textit{O-k10} are highly ranked. 

Nonetheless, there is more to unpack, specifically with the distributions in Fig.~\ref{fig:violin_chart}. First of all, a severe culling of $k=1$ is a risky proposition, regardless of meta-learning targeting strategy. If the remaining predictor is a strong choice for a dataset, the severely reduced search space allows this predictor to be hyperparametrically fine-tuned with greater focus than in any other scenario, meaning that $k=1$ configuration spaces are often considered most beneficial, i.e. ranked 1st. Small predictor search spaces are also more supportive of multi-component pipeline evaluations, e.g. the \textit{L-k1} structure for \textit{abalone} in Table \ref{tab:pipelines}. However, $k=1$ spaces are also frequently the least beneficial, i.e. ranked 18th, especially if the sole predictor is not well-suited for an ML problem. In fact, Fig.~\ref{fig:violin_chart} suggests that AutoML optimisation should actually hedge its bets with an elite tier of predictors, i.e. $k>1$. This is particularly evident for \textit{landmarked} and \textit{oracle} configuration spaces, where the ranking distributions shift substantially towards the top-performing half of the 18 scenarios, i.e. closer to the bottom axis. Of course, these meta-learning gains eventually dissipate for extreme values of $k$, as evidenced by the $k=19$ distributions. After all, a culling scenario with $k=30$ would be equivalent to the standard \textit{avatar} setting.

Further discussion is specific to each type of meta-learning experiment.

\textit{Global leaderboard configuration spaces:}
Figure \ref{fig:ranking_predictors} shows 30 predictors sorted by their average performance-ranking across 11 ML problems, as derived from the meta-knowledge base, additionally displaying how they ranked on each individual dataset. As is evident from this figure, the benefit of searching throughout the top $k$ components of such an ordering is that the top tier of predictors is selected for consistency. Indeed, random forest is a ubiquitous benchmark in Kaggle competitions for that very reason. Likewise, complex ensemble methods, listed in Fig.~\ref{fig:ranking_predictors} as `meta.X', are generally not to be recommended for small optimisation time budgets. Accordingly, the \textit{M-kn} distributions in Fig.~\ref{fig:violin_chart} are relatively unimodal, with their recommendations stably performant for AutoML across all datasets. Additionally, the `all-rounders' suggested by both \textit{M-k1} and \textit{M-k4} seem a good choice for any dataset. On the other hand, without targeting the characteristics of an ML problem, the benefits of versatile predictors are quickly lost to the increased search spaces, as evidenced by \textit{M-k8} and \textit{M-k10}.

\textit{Oracle configuration spaces:}
The \textit{oracle} setting is a theoretical ideal, unlikely to be used in practice. Discounting for the compounded uncertainties in our meta-learning experiments, \textit{O-kn} should be the optimal recommendation procedure for establishing configuration space, given that it leverages what worked previously on the ML problem on hand in a form of direct memory. As already noted, the risk of selecting the wrong predictor, i.e. $k=1$, is evident in the broad distribution of \textit{O-k1}. However, once bets are hedged with a top tier of predictors, i.e. $k>1$, the spaces defined by \textit{O-k4}, \textit{O-k8} and \textit{O-k10} are all very effective culling strategies, with worst-case rankings higher than any alternative. Moreover, as Fig.~\ref{fig:cd_diagram} shows, the average ranking of 5.27 for \textit{O-k4} is more than 7.942 away from \textit{baseline}, i.e. it is critically different and thus a significant result despite the uncertainty inherent to the meta-knowledge base.

\textit{Landmarked configuration spaces:}
More so than either \textit{global leaderboard}
or even \textit{oracle}, the landmarker-based recommendations of configuration space appear significantly bimodal. In effect, they either work well or fail badly. It is possible that the bifurcation is partially due to the meta-learning approach we employed; a refined selection of landmarkers or another method of determining dataset similarity could potentially weight the distributions more towards the high-ranking mode. Regardless, unlike \textit{global leaderboard}, landmarked recommendations remain competitive for a greater range of $k$, possibly because the marginal utility of an extra component in the search space is greater for \textit{L-kn} than \textit{M-kn} due to the predictor being more relevant to the problem on hand. As a result, while increasing values of $k$ diminish the best-case utility of a culled space, i.e. the maximum performance ranking, the average actually increases for a while, such that \textit{L-k10} proves to be the second-most reliable recommendation scheme of all 18 scenarios. In fact, according to the statistical tests underlying the CD diagram in Fig.~\ref{fig:cd_diagram}, the average rank for \textit{L-k10} of 6.68 is more than 7.942 away from the \textit{baseline} rank of 15.09, i.e. a critical difference. This result is significant and strongly supports the validity of using meta-knowledge in an opportunistic fashion to boost AutoML optimisation processes.

Finally, we emphasise that these results are intrinsically dependent on the time budget used for ML pipeline composition/optimisation, which, in these experiments, has been two hours per SMAC run. The consequences of this choice can be counter-intuitive. For instance, settings \textit{r30} and \textit{O-k1} are effectively identical, except that the former previously had 30 hours to select its best pipeline, while the latter only had two. So, it would seem that \textit{r30} has an advantage. Instead, the violin plots in Fig.~\ref{fig:violin_chart} show that \textit{O-k1} is superior. Admittedly, as Table~\ref{tab:pipelines} implies, the extended exploration time does allow \textit{r30} to recommend far more complex ML pipelines in its culled space than \textit{O-k1}. However, a one-component pipeline can have its hyperparameters optimised far more effectively in two hours. In fact, two hours may not be enough to even train the alternative, let alone iterate through its hyperparameter configurations.

On that topic, we note the following. With infinite time, an encompassing search space of $k=30$ will always provide the best solutions that $k<30$ risks missing out on. At the other extreme, with negligible time, the only feasible option is $k=1$, requiring a desperate choice of predictor that is hopefully informed by strong meta-knowledge. For any other practical choice of runtime, there will be a `sweet spot' for tier size $k$. In our particular experiments, and based on the average rankings of cull strategies, these were denoted by \textit{M-k4}, \textit{L-k10} and \textit{O-k4}. Such results may have a large degree of uncertainty, but the underlying principle is clear: although imperfect, even weakly-biasing meta-knowledge can boost the search for ML solutions that is subject to strict time limits.

\section{Conclusion}
\label{sec:conclusion}

In this study, we have investigated whether the routine process of AutoML optimisation, previously applied to a collection of datasets, can provide any useful information to support model search in the future. Specifically, we opportunistically harvested numerous evaluations of ML pipelines, substantial but non-exhaustive, to produce a meta-knowledge base of 30 predictors and their ranked performance on each of 20 datasets. We then ran a series of experiments with AutoML package AutoWeka4MCPS and its pipeline composition/optimisation algorithm SMAC, in which the solution search space for a target ML problem was culled to varying sets of predictors informed by the meta-knowledge base. These recommendations could be based on how predictors performed overall, how they performed on the most similar dataset to the one on hand, or how they performed on the dataset itself in a previous run. Dataset similarity, where relevant, was determined by the method of relative landmarking.

We found that, despite the intended unreliability of the meta-knowledge base, meta-learning does, as a generalisation, improve the outcome of SMAC. Moreover, AutoML solution search appears to do better the more relevant the meta-knowledge is to a dataset on hand. The impact of landmarker-based search-space recommendation even proved critically different from our baseline strategy, although future experiments will be required to further establish the statistics of these results.

Ultimately, we find that our studied form of opportunistic meta-knowledge, compiled with a minimal level of thoroughness, is risky to depend on when selecting the best predictor for a dataset; its optimised performance is frequently `all or nothing'. In contrast, the meta-knowledge proves much more useful in culling away the worst performers, so as to leave behind a top tier of potential ML models, the optimal size of which depends on the runtime available for optimisation. In effect, our research suggests that AutoML should seek a risk-averse balance that ensures promising candidates are not disregarded, while also dedicating enough time to properly trial them.

\bibliographystyle{IEEEtran}
\bibliography{references}

\end{document}